\documentclass[a4paper]{article}
\usepackage[it]{idsiatechrep}
\usepackage{times}
\usepackage[english]{babel}
\usepackage{graphicx}
\usepackage{amsmath, amsthm, amssymb}
\usepackage{algorithm}
\usepackage{algorithmic}
\usepackage{url}

\newcommand{\x}{\mathbf{x}}

\newcommand{\share}{\mathbf{s}}

\newcommand{\algset}{\mathcal{A}}
\newcommand{\taskset}{\mathcal{B}}

\newcommand{\talloc}{\mbox{TA}}
\newcommand{\tallocset}{\mathcal{T}}

\newcommand{\nalgs}{K}
\newcommand{\ntasks}{M}
\newcommand{\nexps}{N}
\newcommand{\narms}{N}%{K}
\newcommand{\ntrials}{M}
\newcommand{\nruns}{20}

\newcommand{\lmax}{\mathcal{L}}
\newcommand{\Loss}{L}
\newcommand{\loss}{l}
\newcommand{\Lest}{\tilde{\Loss}}
\newcommand{\lest}{\tilde{\loss}}
\newcommand{\bps}{E}
\newcommand{\lalg}{\loss_{\bps}}
\newcommand{\Lalg}{\Loss_{\bps}}
\newcommand{\epochLbest}{r}
\newcommand{\epochlmax}{u}
\newcommand{\lastepochlmax}{U}
\newcommand{\vecpbandit}{\mathbf{p}}
\newcommand{\BPS}{{\sc BPS}}

\newcommand{\GambleTA}{{\sc GambleTA}}
\newcommand{\GambleTAThree}{{\sc GambleTA3}}
\newcommand{\GambleTAFour}{{\sc GambleTA4}}
\newcommand{\ExpFour}{{\sc Exp4}}

\newcommand{\ExpThreeLight}{{\sc Exp3Light}}
\newcommand{\ExpThreeLightA}{{\sc Exp3Light-A}}
\newcommand{\Oracle}{{\sc Oracle}}
\newcommand{\Uniform}{{\sc Uniform}}

\newcommand{\Expval}[1]{E\{#1\}}
\newcommand{\ceil}[1]{\lceil#1\rceil}

\newtheorem{theorem}{Theorem}

\title{Algorithm Selection as a Bandit Problem with Unbounded Losses}
\author{ Matteo Gagliolo \and 
J\"{u}rgen Schmidhuber} 
\date{July 9, 2008}
\reportnumber{07-08}
\reportaddnote{Both authors are also affiliated with the University of Lugano, Faculty of Informatics (Via Buffi 13, 6904 Lugano, Switzerland). J. Schmidhuber is also affiliated with  TU Munich  (Boltzmannstr. 3,  85748 Garching, M\"{u}nchen, Germany). \\
This work was supported by the Hasler
foundation with grant n.~$2244$.}

\begin{document}
\makecover
\maketitle
\begin{abstract} 
Algorithm selection is typically based on models of algorithm performance, 
learned during a separate \emph{offline} training sequence, which can be 
prohibitively expensive. In recent work, we adopted an \emph{online} approach,
 in which a performance model is iteratively updated and used to guide selection
on a sequence of problem instances. The resulting \emph{exploration-exploitation}
trade-off was represented as a bandit problem with expert advice, using an 
existing solver for this game, but this required the setting of an arbitrary 
bound on algorithm runtimes, thus invalidating the optimal regret of the solver. 
In this paper, we propose a simpler framework for representing algorithm selection
 as a bandit problem, with partial information, and an \emph{unknown} bound on losses.
 We adapt an existing solver to this game, proving a bound on its expected regret, 
which holds also for the resulting algorithm selection technique. We present 
preliminary experiments with a set of SAT solvers on a mixed SAT-UNSAT benchmark.
\end{abstract}

\section{Introduction}
\label{intro}

Decades of research in the fields of Machine Learning and Artificial
Intelligence brought us a variety of alternative algorithms
for solving many kinds of problems. Algorithms often
display variability in performance quality, and computational
cost, depending on the particular problem instance being solved:
in other words, there is no single ``best'' algorithm.
While a ``trial and error'' approach is still the most
popular, attempts to automate algorithm selection are not new 
 \cite{rice76}, and have grown to form a consistent
and dynamic field of research in the area of \emph{Meta-Learning} \cite{Vilalta:02airev}.
Many selection methods follow an \emph{offline} learning scheme,
in which the availability of a  large training set of performance
data for the different algorithms is assumed.
This data is used to learn a model that %can 
maps (\emph{problem}, \emph{algorithm}) pairs
to expected performance, or to some probability distribution
on performance.
The model is later used to select and run, for each new problem instance,
only the algorithm that is expected to give the best results.
While this approach might sound reasonable, it actually ignores
the computational cost of the initial training phase:
collecting a representative sample of performance data has to be done via  
solving a set of training problem instances, and each instance is
solved repeatedly, at least once for each of the available algorithms, 
or more if the algorithms are randomized. Furthermore, these training
instances are assumed to be representative of future ones, as the model
is not updated after training. 

In other words,
there is an obvious trade-off 
between the \emph{exploration} of algorithm performances
on different problem instances, aimed at learning the model,
and the \emph{exploitation} of the best algorithm/problem combinations,
based on the model's predictions. 
This trade-off is typically ignored 
in offline algorithm selection, and the size of the training set 
is chosen heuristically.
In our previous work \cite{Gagliolo:04ecml,Gagliolo:05icann,Gagliolo:06amai}, we
have kept an \emph{online} view of algorithm selection, in which the
only input available to the meta-learner is a set of algorithms, of 
unknown performance, and a sequence of problem instances that have to be solved.
Rather than artificially subdividing the problem set into
a training and a test set, we iteratively update the model 
each time an instance is solved, and use it to guide algorithm
selection on the next instance.

Bandit problems \cite{Auer05SIAM} offer a solid theoretical framework for dealing with
the exploration-exploitation trade-off in an online setting. 
One important obstacle to the straightforward
application of a bandit problem solver to algorithm
selection is that most existing solvers assume
 a bound on losses to be available beforehand.
In \cite{Gagliolo:07ijcai,Gagliolo:06amai} we dealt with this
issue heuristically, fixing the bound in advance.
In this paper, we introduce a modification of an existing
bandit problem solver \cite{cesa-bianchi05improved},
which allows it to deal with an unknown bound on losses,
while retaining a bound on the expected regret.
This allows us to propose a simpler version of 
the algorithm selection framework \GambleTA, originally introduced in 
\cite{Gagliolo:06amai}. The result is a 
parameterless online algorithm selection method, 
 the first, to our knowledge, with a provable upper bound on regret.

The rest of the paper is organized as follows.
Section \ref{previous} describes a tentative taxonomy
of algorithm selection methods, along with a few
examples from literature. Section \ref{banditas}
presents our framework for representing
 algorithm selection as a bandit problem, discussing
the introduction of a higher level of selection among
different algorithm selection techniques (\emph{time allocators}). 
Section \ref{exp3light} introduces the modified
bandit problem solver for unbounded loss games, 
along with its bound on regret.
Section \ref{experiments} describes  
experiments with SAT solvers.
Section \ref{conclusions} concludes the paper.

\section{Related work}
\label{previous}

In general terms, algorithm selection can be defined as the process
of allocating computational resources to a set of alternative algorithms,
in order to improve some measure of performance on a set of problem
instances. Note that this definition includes parameter selection: 
the algorithm set can contain multiple copies of 
a same algorithm, differing in their parameter settings; or
even identical randomized algorithms differing only in their random seeds. 
 Algorithm selection techniques can be further described according to different orthogonal features:

{\bf Decision vs. optimisation problems.} A first distinction needs to be made
among \emph{decision} problems, where a binary criterion for recognizing 
a solution is available; and  \emph{optimisation}
problems, where different levels of solution quality can be attained,
measured by an \emph{objective} function \cite{Hoos2000a}.
Literature on algorithm selection is often focused on
one of these two classes of problems. The selection is normally
aimed at  minimizing solution time for decision problems;  
and at maximizing performance quality, or improving
some speed-quality trade-off, for optimisation problems.\\
{\bf Per set vs. per instance selection.} The selection among different 
algorithms can be performed once for an entire set of
problem instances (\emph{per set} selection, following \cite{HutHam05});
or repeated for each instance (\emph{per instance} selection).\\
{\bf Static vs. dynamic selection.} A further independent
distinction \cite{Petrik05} can be made among \emph{static} algorithm selection,
in which allocation of resources
precedes algorithm execution; and  \emph{dynamic}, or
\emph{reactive}, algorithm selection,
in which the allocation can be adapted during algorithm execution.\\
{\bf Oblivious vs. non-oblivious selection.} In \emph{oblivious} techniques,
algorithm selection is performed from scratch for each problem instance;
in \emph{non-oblivious} techniques, there is some knowledge transfer
across subsequent problem instances, usually in the form of 
a \emph{model} of algorithm performance.\\
{\bf Off-line vs. online learning.} Non-oblivious techniques
can be further distinguished as \emph{offline} or \emph{batch} learning techniques, where 
a separate training phase is performed, after which the selection criteria 
are kept fixed; and \emph{online}
techniques, where the criteria can be updated every time an instance is solved.

 A seminal paper in the field of algorithm selection 
 is \cite{rice76}, in which offline, per instance
 selection is first proposed,
for both decision and optimisation problems.
 More recently, similar concepts have been proposed,
with different terminology (algorithm \emph{recommendation}, \emph{ranking}, \emph{model selection}),
 in the \emph{Meta-Learning} community \cite{metal,Vilalta:02airev,mach:Giraud-Carrier+Vilalta+Brazdil:2004}. 
Research in this field usually  deals with optimisation problems,
and is focused on maximizing solution quality, without taking into  account the computational aspect.
Work on \emph{Empirical Hardness Models} \cite{LeyNudSho02,Nudelman04sat}
 is instead applied to decision problems, and focuses on
obtaining accurate models of runtime performance, conditioned
on numerous features of the problem instances,
as well as on parameters of the solvers \cite{HutHam05}. The models
are used to perform algorithm selection
on a per instance basis, and are learned offline: online selection
is advocated in \cite{HutHam05}. 
Literature on algorithm portfolios \cite{Huberman:97,gomes-selman:2001a,Petrik06aimath}  is usually focused on choice
criteria for building the set of candidate solvers, such that
their areas of good performance do not overlap, and optimal
static allocation of computational resources among elements
of the portfolio.

A number of interesting dynamic exceptions to the static selection
paradigm have been proposed recently. 
In \cite{Kautz:02aaai}, algorithm performance modeling is 
based on the behavior of the candidate algorithms during a predefined
amount of time, called the {\em observational horizon}, and
dynamic context-sensitive restart policies for SAT solvers are presented.
In both cases, the model is learned offline.
In a Reinforcement Learning \cite{Sutton:98} setting,
algorithm selection can be formulated as a
Markov Decision Process: in \cite{ll-asrl-00},  
the algorithm set includes sequences of recursive algorithms,
formed dynamically at run-time solving a sequential decision problem,
and a variation of Q-learning is used to find a dynamic algorithm selection policy;
the resulting technique is per instance, dynamic and online.
In \cite{Petrik05},
 a set of deterministic algorithms is considered, and, 
under some limitations, static and dynamic schedules are obtained, 
based on dynamic programming. In both cases, the method
presented is per set, offline.

An approach based on runtime distributions can be found in 
\cite{finkelstein02optimal-i,finkelstein02optimal-s},
for parallel independent processes and shared resources respectively.
The runtime distributions are assumed to be known, 
and the expected value of a cost function, accounting for 
both wall-clock time and resources usage, is minimized. 
A dynamic schedule is evaluated offline,
using a branch-and-bound algorithm to find the optimal one 
in a tree of possible schedules.
Examples of allocation to two processes are presented 
with artificially generated runtimes, and a real
Latin square solver. Unfortunately, the computational complexity 
of the tree search grows exponentially in the number of processes.

``Low-knowledge'' oblivious approaches can be found in \cite{Beck:04CPAIOR,Carchrae:05ci},
in which various simple indicators of current solution improvement are used 
for algorithm selection, in order to achieve the best solution quality
within a given time contract. 
In \cite{Carchrae:05ci}, the selection 
process is dynamic: machine time shares are based
on a recency-weighted average of performance improvements.
 We adopted a similar approach in \cite{Gagliolo:04ecml}, 
where we considered algorithms with a scalar state, 
that had to reach a target value. 
The time to solution was estimated based on a shifting-window 
linear extrapolation of the learning curves.

For optimisation problems, if selection is only aimed at maximizing solution 
quality,  the same problem instance can be solved multiple times,
 keeping only the best  solution.
In this case, algorithm selection  can be represented as a
\emph{Max} $K$-armed bandit problem,
a variant of the game in which the reward attributed to each
arm is the maximum payoff on a set of rounds. 
Solvers for this game are used in \cite{Cicirello:05aaai,Streeter:06aaai}
to implement oblivious per instance selection from a set of multi-start 
optimisation techniques: each problem is treated independently,
and multiple runs of the available solvers are allocated, to maximize
performance quality.
Further references can be found in \cite{Gagliolo:06amai}.

\section{Algorithm selection as a bandit problem}
\label{banditas}

In its most basic form \cite{Robbins:52}, the \emph{multi-armed bandit} problem 
is faced by a gambler, playing a sequence of trials against an $\narms$-armed slot machine. 
At each trial, the gambler chooses one
of the available arms, whose losses are randomly generated 
from different \emph{stationary} distributions.
 The gambler incurs in the corresponding 
loss, and, in the \emph{full information} game,
 she can observe the losses that would have been paid pulling
any of the other arms. A more optimistic formulation can be made in terms of positive rewards.
The aim of the game is to minimize the \emph{regret}, defined as the difference
between the  cumulative loss of the gambler,  and the one of the best arm. 
A bandit problem solver (\BPS)
can be described as a mapping from the history of the observed
losses $l_j$ for each arm $j$, to a probability distribution $\mathbf{p}=(p_1,...,p_{\narms})$,
from which the choice for the successive trial will be picked.

More recently, the original restricting assumptions have been 
progressively relaxed, allowing for \emph{non-stationary} loss distributions,
\emph{partial} information (only the loss for the pulled arm is observed),
and \emph{adversarial} bandits that can set their losses in order
to deceive the player.
In \cite{Auer:1995:GRC,Auer05SIAM}, 
a reward game is considered, and
no statistical assumptions are made about
the process generating the rewards, which are allowed to be an
arbitrary function of the entire history of the game
(\emph{non-oblivious} adversarial setting).
Based on these pessimistic hypotheses, the authors describe
probabilistic gambling strategies for the full 
and the partial information games.

Let us now see how to represent algorithm selection for \emph{decision}
problems as a bandit problem, with the aim of minimizing solution time.
 Consider  a sequence $\taskset=\{b_1,\ldots,b_{\ntasks}\}$ 
of $\ntasks$ instances of a decision problem, 
 for which we want to minimize solution time,
and a set of $\nalgs$ algorithms $\algset=\{a_1,\ldots,a_{\nalgs}\}$, 
such that each $b_m$ can be solved by each $a_k$.
It is straightforward to describe 
static algorithm selection
 in a multi-armed bandit setting,
where ``pick arm $k$'' means ``run algorithm $a_k$ on next problem instance''.
 Runtimes $t_k$ can be viewed as losses,
generated by a rather complex mechanism,
i.e., the algorithms $a_k$ themselves, running on the current problem.
The information is partial, as the runtime for other algorithms is 
not  available, unless we decide to solve the same 
problem instance again. 
 In a worst case scenario one can receive a
''deceptive'' problem sequence, starting with 
problem instances on which the performance
of the algorithms is misleading, 
so this bandit problem should be considered  
adversarial. 
 As \BPS\  typically minimize the regret with respect to a single
arm, this approach would allow to implement \emph{per set} selection,
of the overall best algorithm.
An example can be found in \cite{Gagliolo:07ijcai}, 
where we presented an online method for
learning a per set estimate of an optimal restart strategy.

Unfortunately, per set selection is only profitable
if one of the algorithms
dominates the others on all problem instances.
This is usually not the case: 
it is often observed in practice
that different algorithms perform better on 
different problem instances. In this situation,
a per instance selection scheme, which can 
take a different decision 
for each problem instance,
can have a great advantage. 

One possible way of exploiting the nice theoretical
properties of a \BPS\  in the context of algorithm selection,
 while allowing for the improvement in performance
of per instance selection, is to use the \BPS\  at 
an upper level, to select among alternative
algorithm selection techniques. 
Consider again the algorithm selection
problem represented by $\taskset$ and $\algset$.
Introduce a set of $\nexps$  \emph{time allocators} ($\talloc_j$) 
\cite{Gagliolo:04ecml,Gagliolo:06amai}.
Each $\talloc_j$ can be an arbitrary function, mapping the 
current history of collected performance data
for each $a_k$, to a share $\share^{(j)}\in[0,1]^{\nalgs}$,
with $\sum_{k=1}^{\nalgs}s_k=1$.
A $\talloc$ is used to solve a given problem instance
executing all algorithms in $\algset$ in parallel,
on a single machine, whose computational resources 
are allocated to each $a_k$
proportionally to the corresponding $s_k$, 
such that for any portion of time spent $t$, $s_kt$ is used
by $a_k$, as in a \emph{static} algorithm portfolio 
\cite{Huberman:97}.
The runtime before a solution is found is then 
$\min_k\{t_k/s_k\}$, $t_k$ being the runtime of algorithm $a_k$.

A trivial example of a $\talloc$ is the \emph{uniform} time
allocator, assigning a constant $\share=(1/K,...,1/K)$. 
Single algorithm selection can be represented
in this framework by setting a single $s_k$ to $1$.
Dynamic allocators will produce a time-varying share $\share(t)$. 
In previous work, we presented examples of heuristic oblivious
\cite{Gagliolo:04ecml} and non-oblivious \cite{Gagliolo:05icann}
 allocators; 
more sound $\talloc$s are proposed in \cite{Gagliolo:06amai},
based on non-parametric models of the runtime distribution
of the algorithms, which are used to minimize
the expected value of solution time, 
or a \emph{quantile} of this quantity,
or to maximize solution probability
within a give time \emph{contract}.

At this higher level, one can use a \BPS\ 
 to select among different time allocators,
$\talloc_j, \talloc_2 \ldots$,
working on a same algorithm set $\algset$. In this case,
``pick arm $j$'' means ``use time allocator $\talloc_j$ 
on $\algset$ to solve next problem instance''. In the long term,
the \BPS\  would allow to select, on a \emph{per set} basis,
 the $\talloc_j$ that is best
at allocating time to algorithms in $\algset$ on a \emph{per instance} basis. 
The resulting ``Gambling'' Time Allocator (\GambleTA) is described in Alg.~\ref{gambleta}. 

\begin{algorithm}
\caption{\GambleTA$(\algset,\tallocset,BPS)$ Gambling Time Allocator. }
\label{gambleta}
\begin{algorithmic}
\STATE Algorithm set $\algset$ with $\nalgs$ algorithms; 
\STATE A set $\tallocset$ of  $\nexps$ time allocators $\talloc_j$;
\STATE A bandit problem solver \BPS\  
\STATE $\ntasks$ problem instances.
\STATE
\STATE initialize \BPS$(\nexps,\ntasks)$ 
\FOR{each problem $b_i, i=1,\ldots,\ntasks$}
\STATE pick time allocator $I(i)=j$ with probability $p_j(i)$ from \BPS.
\STATE solve problem $b_i$ using $\talloc_{I}$ on $\algset$
\STATE incur loss  $\loss_{I(i)}=\min_k\{t_k(i)/s_k^{(I)}(i)\}$
\STATE update \BPS\ 
\ENDFOR
\end{algorithmic}
\end{algorithm}

If \BPS\  allows for non-stationary arms, it can 
also deal with time allocators that are \emph{learning}
to allocate time. This is actually the original motivation
for adopting this two-level selection scheme,
as it allows to combine in a principled way
the exploration of algorithm behavior,
which can be represented by the uniform time allocator,
and the exploitation of this information by %: this is precisely the situation
a model-based allocator, whose model is being
learned online, based on results on the sequence
of problems met so far. If more time allocators are available,
they can be made to compete, using the \BPS\  to explore
their performances. 
Another interesting
feature of this selection scheme is that 
the initial requirement that each algorithm 
should be capable of solving each problem 
can be relaxed, requiring instead that at least
one of the $a_k$ can solve a given $b_m$,
and that each $\talloc_j$ can solve
each $b_m$: this can be ensured in practice
by imposing a $s_k>0$ for all $a_k$. 
This allows
to use interesting combinations
of complete and incomplete solvers
in $\algset$ (see Sect.~\ref{experiments}).
Note that any bound on the regret of the \BPS\  
will determine a bound on the regret of \GambleTA\  
with respect to the best time allocator. 
Nothing can be said
about the performance w.r.t. the best algorithm.
In a worst-case setting, if none of the time allocator
 is effective, a bound can still
be obtained by including the uniform share in
the set of $\talloc$s. In practice, though, per-instance selection
can be much more efficient than uniform allocation,
and the literature is full of examples of time allocators
which eventually converge to a good performance.

The original version of \GambleTA\  (\GambleTAFour\ in the following)
 \cite{Gagliolo:06amai} was based on 
a more complex alternative, inspired by the 
bandit problem with \emph{expert} advice, 
as described in \cite{Auer:1995:GRC,Auer05SIAM}.
In that setting,
two games are going on in parallel: at a lower level,
a partial information game is played, based on the 
probability distribution obtained \emph{mixing}
the advice of different \emph{experts}, represented as
probability distributions on the $K$ arms.
 The experts can be arbitrary functions
of the history of observed rewards, 
and give a different advice for each trial.
At a higher level, a \emph{full information} game
is played, with the $\nexps$ experts playing the roles 
of the different arms. The probability distribution
$\vecpbandit$ at this level is not used to pick a single
expert, but to \emph{mix} their advices, in order to generate
the distribution  for the lower level arms.
In \GambleTAFour, the time allocators play the role
of the experts, each suggesting a different $\share$,
on a per instance basis; and the arms of the lower
level game are the $\nalgs$ algorithms,
to be run in parallel with the mixture share.
\ExpFour\   \cite{Auer:1995:GRC,Auer05SIAM} is used as the \BPS. 
Unfortunately, the bounds for \ExpFour\  cannot be
extended to \GambleTAFour\  in a straightforward manner, 
as the loss function itself is not convex;
moreover, \ExpFour\  cannot deal with unbounded
losses, so we had to adopt an heuristic reward
attribution instead of using the plain runtimes.

A common issue of the above approaches
is the difficulty of setting reasonable upper bounds on the 
time required by the algorithms. This renders 
a straightforward application of most \BPS\  problematic,
as a known bound on losses is usually assumed, and used
to tune parameters of the solver.
Underestimating this bound can invalidate the bounds on regret,
 while overestimating it can  produce
an excessively ''cautious'' algorithm, with a poor performance.
Setting in advance a good bound is particularly difficult
when dealing with algorithm runtimes, which can easily
exhibit variations of several order of magnitudes
among different problem instances, or even among different
runs on a same instance \cite{Gomes:2000jar}.

Some interesting results regarding games with \emph{unbounded}
losses have recently been obtained. 
In \cite{cesa-bianchi05improved,cesa-bianchi07improved}, the authors 
consider a full information game, and provide two algorithms 
which can adapt to unknown bounds on signed rewards.
Based on this work, \cite{allenberg06hannan} provide
a Hannan consistent algorithm for losses whose
bound grows in the number of trials $i$  with a known rate
 $i^{\nu}$, $\nu<1/2$. This latter
hypothesis does not fit well our situation, as
we would like to avoid any restriction on the
sequence of problems:  
a very hard instance can be met first, 
followed by an easy one. In this sense, the 
hypothesis of a constant, but unknown,
bound is more suited.
In \cite{cesa-bianchi05improved}, Cesa-Bianchi \emph{et al.} also introduce an
 algorithm for loss games with partial information (\ExpThreeLight),
which requires losses to be bound, and is 
particularly effective when the cumulative loss of the best arm 
is small. 
 In the next section we introduce a variation
of this algorithm that allows it to deal with an unknown bound
on  losses.

\section{An algorithm for games with an unknown bound on losses}
\label{exp3light}

Here and in the following, we consider a partial information game
with $\narms$ arms, and $\ntrials$ trials;
 an index $(i)$ indicates the value of 
a quantity used or observed at trial 
$i\in\{1,\ldots,\ntrials\}$; $j$ indicate quantities related to the $j$-th arm,
 $j\in\{1,\ldots,\narms\}$; 
index $\bps$ refers to the loss incurred
by the bandit problem solver, and $I(i)$ indicates
the arm chosen at trial $(i)$, so it is
a discrete random variable with value in $\{1,\ldots,\narms\}$;
$\epochLbest$,
$\epochlmax$ will represent quantities related
to an \emph{epoch} of the game, which consists of a sequence
of $0$ or more consecutive trials; $\log$ with no index is the natural
logarithm.

\ExpThreeLight\  \cite[Sec. 4]{cesa-bianchi05improved}
is a solver for the bandit loss game with partial information.
It is a modified version of the weighted majority algorithm
 \cite{Littlestone94weighted}, in which the cumulative losses for each arm 
are obtained through an unbiased estimate\footnote{
For a given round, and a given arm with loss $l$ and 
pull probability $p$, the estimated loss $\lest$ is $l/p$ if the arm is pulled,
 $0$ otherwise. This estimate is unbiased in the sense that its expected value,
with respect to the process extracting the arm to be pulled, equals
the actual value of the loss: $\Expval{\lest}= pl/p + (1-p)0=l$.}. 
The game consists of a sequence of epochs $\epochLbest=0,1,\ldots$:
in each epoch, the probability distribution over 
 the arms is updated, proportional
to $\exp{(-\eta_{\epochLbest}\Lest_j)}$, $\Lest_j$ being the current
unbiased estimate 
of the cumulative loss. Assuming an upper bound $4^\epochLbest$ on the 
smallest loss estimate, $\eta_{\epochLbest}$ is set as:
\begin{equation}
\label{eqetar}
\eta_{\epochLbest}=\sqrt{\frac{2(\log \narms +\narms\log\ntrials)}{(\narms 4^{\epochLbest})}}
\end{equation}
When this bound is first trespassed, a new epoch starts and 
$\epochLbest$ and $\eta_{\epochLbest}$ are updated accordingly. 

\begin{algorithm}
\caption{\ExpThreeLight$(\narms,\ntrials,\lmax)$ A solver for bandit problems with partial information and 
a known bound $\lmax$ on losses.
 }
\label{Exp3Light}
\begin{algorithmic}
\STATE  $\narms$ arms, $\ntrials$ trials 
\STATE losses $\loss_j(i)\in[0,\lmax]$ $\forall$ $i=1,...,\ntrials$, $j=1,\ldots,\narms$
\STATE initialize epoch $\epochLbest=0$, $\Lalg=0$, $\Lest_j(0)=0$.
\STATE initialize $\eta_{\epochLbest}$ according to (\ref{eqetar})
\FOR{each trial $i=1,...,\ntrials$}
\STATE set $p_j(i)\propto \exp(-\eta_{\epochLbest}\Lest_j(i-1)/\lmax)$, $\sum_{j=1}^{\narms}p_j(i)=1$.
\STATE pick arm $I(i)=j$ with probability $p_j(i)$.
\STATE incur loss $\lalg(i)=\loss_{I(i)}(i)$.
\STATE evaluate unbiased loss estimates: 
\STATE $\lest_{I(i)}(i)=\loss_{I(i)}(i)/p_{I(i)}(i)$, $\lest_j=0$ for $j\neq I(i)$
\STATE update cumulative losses: 
\STATE $\Lalg(i)=\Lalg(i-1)+\lalg(i)$, 
\STATE $\Lest_j(i)=\Lest_j(i-1)+\lest_j(i)$, for $j=1,\ldots,\narms$ 
\STATE $\Lest^{*}(i)=min_j \Lest_j(i)$.
\IF{$ (\Lest^{*}(i)/\lmax) > 4^{\epochLbest}$}
\STATE start next epoch $\epochLbest=\ceil{\log_4 (\Lest^{*}(i)/\lmax)}$
\STATE update $\eta_{\epochLbest}$ according to (\ref{eqetar})      
\ENDIF
\ENDFOR
\end{algorithmic}
\end{algorithm}

The original algorithm assumes losses in $[0,1]$.
We first consider a game with a known finite bound $\lmax$ on 
losses, and introduce a slightly modified version
of \ExpThreeLight\  (Algorithm \ref{Exp3Light}), 
obtained simply dividing all losses by $\lmax$.
Based on Theorem 5 from \cite{cesa-bianchi05improved}, 
it is easy to prove the following

\begin{theorem}
\label{thExp3Light}
If $\Loss^{*}(\ntrials)$ is the loss of the best arm after $\ntrials$ trials, 
and $\Lalg(\ntrials)=\sum_{i=1}^{\ntrials}\loss_{I(i)}(i)$ is the loss of 
\ExpThreeLight$(\narms,\ntrials,\lmax)$, the expected value of its regret 
is bounded as:

\begin{eqnarray}
\label{eqExp3lightBoundLmax}
&&\Expval{\Lalg(\ntrials)}-\Loss^*(\ntrials)  \\
&\leq&2\sqrt{6\lmax(\log \narms +\narms\log\ntrials)\narms\Loss^*(\ntrials)}\nonumber\\
&+&\lmax[2\sqrt{2\lmax(\log \narms +\narms\log\ntrials)\narms}\nonumber\\
&+&(2\narms+1)(1+\log_4(3\ntrials+1))]\nonumber
\end{eqnarray}

\end{theorem}

The proof is trivial, and is given in the appendix.

We now introduce a simple variation of Algorithm \ref{Exp3Light}
 which does not require the knowledge
of the bound $\lmax$ on losses, and uses
Algorithm \ref{Exp3Light} as a subroutine.
 \ExpThreeLightA\  (Algorithm \ref{Exp3LightA}) is inspired by the doubling trick
used in \cite{cesa-bianchi05improved} for a \emph{full} information
game with unknown bound on losses. The game is again organized in a sequence
of epochs $\epochlmax=0,1,\ldots$: in each epoch, Algorithm \ref{Exp3Light}
is \emph{restarted} using a bound $\lmax_{\epochlmax}=2^{\epochlmax}$; a new epoch is started
with the appropriate $\epochlmax$ whenever a loss larger 
than the current $\lmax_{\epochlmax}$ is observed.

\begin{algorithm}
\caption{\ExpThreeLightA$(\narms,\ntrials)$ 
A solver for bandit problems with partial information and an unknown 
 (but finite) bound on losses.}
\label{Exp3LightA}
\begin{algorithmic}
\STATE  $\narms$ arms, $\ntrials$ trials, 
\STATE losses $\loss_j(i)\in[0,\lmax]$ $\forall$ $i=1,...,\ntrials$,$j=1,\ldots,\narms$
\STATE {\bf unknown} $\lmax< \infty$ 
\STATE initialize epoch $\epochlmax=0$, \ExpThreeLight$(\narms,\ntrials,2^{\epochlmax})$
\FOR{each trial $i=1,...,\ntrials$}
\STATE pick arm $I(i)=j$ with probability $p_j(i)$ from \ExpThreeLight.
\STATE incur loss $\lalg(i)=\loss_{I(i)}(i)$.
\IF{$ \loss_{I(i)}(i) > 2^{\epochlmax}$}
\STATE start next epoch $\epochlmax=\ceil{\log_2\loss_{I(i)}(i) }$
\STATE {\bf restart} \ExpThreeLight$(\narms,\ntrials-i,2^{\epochlmax})$     
\ENDIF
\ENDFOR
\end{algorithmic}
\end{algorithm}

\begin{theorem}
\label{thExp3LightA}

If $\Loss^*(\ntrials)$ is the loss of the best arm after $\ntrials$ trials, 
and $\lmax<\infty$ is the unknown bound on losses,
the expected value of the regret of \ExpThreeLightA$(\narms,\ntrials)$
is bounded as:

\begin{eqnarray}
\label{eqExp3lightABound}
&&\Expval{\Lalg(\ntrials)}
-\Loss^{*}(\ntrials)\leq\\
&&4\sqrt{3\ceil{\log_2\lmax}\lmax(\log \narms +\narms\log\ntrials)\narms\Loss^{*}(\ntrials)}\nonumber\\
&+&2\ceil{\log_2\lmax}\lmax[ \sqrt{4\lmax(\log \narms +\narms\log\ntrials)\narms}\nonumber\\
&+&(2\narms+1)(1+\log_4(3\ntrials+1))+2]\nonumber
\end{eqnarray}

\end{theorem}

The proof is given in the appendix. The regret obtained 
by \ExpThreeLightA\  is $O(\sqrt{\lmax\narms\log\ntrials\Loss^{*}(\ntrials)})$,
 which can be useful in a situation in which $\lmax$
is high but $\Loss^{*}$ is relatively small, as we
expect in our time allocation setting if the algorithms
exhibit huge variations in runtime, but at least 
one of the $\talloc$s eventually converges to a good
performance.
We can then use \ExpThreeLightA\  as a \BPS\  for selecting
among different time allocators in \GambleTA\  (Algorithm \ref{gambleta}).

\section{Experiments}
\label{experiments}

The set of time allocator used in the following experiments is the same as in  \cite{Gagliolo:06amai},
and includes the uniform allocator, along with nine other \emph{dynamic}
allocators, optimizing different quantiles of runtime, based on a nonparametric
model of the runtime distribution that is updated after each problem is solved.
We first briefly describe these time allocators, inviting the reader to refer to  \cite{Gagliolo:06amai}
for further details and a deeper discussion.
A separate model $F_k(t|\x)$, conditioned on features $\x$ of the problem instance,
is used for each algorithm $a_k$. Based on these models, the runtime distribution
for the whole algorithm portfolio $\algset$ can be evaluated for an arbitrary share
 $\share\in[0,1]^{\nalgs}$, with $\sum_{k=1}^{\nalgs}s_k=1$, as %(see  \cite{Gagliolo:06amai})
\begin{equation}
\label{apcdf}
  F_{\algset,\share}(t) =  1-\prod_{k=1}^{\nalgs} [1-F_k(s_kt)].
\end{equation}

Eq.~(\ref{apcdf}) can be used to evaluate a quantile \mbox{$t_{\algset,\share}(\alpha)=F_{\algset,\share}^{-1}(\alpha)$}
for a given solution probability $\alpha$. Fixing this value, 
time is allocated using the share that minimizes the quantile
\begin{equation}
\label{quantileta}
\share=\arg\min_{\share} F_{\algset,\share}^{-1}(\alpha).
\end{equation}
Compared to minimizing expected runtime, this time allocator has the advantage of being
applicable even when the runtime distributions are improper, i.~e. $F(\infty)<1$, as
in the case of incomplete solvers. 
A \emph{dynamic} version of this time allocator is obtained updating the share value
periodically, conditioning each $F_k$ on the time spent so far by the corresponding $a_k$.

Rather than fixing an arbitrary $\alpha$, we used 
nine different instances of this time allocator, with $\alpha$ ranging from $0.1$ to
$0.9$, in addition to the uniform allocator, and let the \BPS\   select the best one. 

We present experiments for  the algorithm selection scenario from \cite{Gagliolo:06amai},
in which a local search and a complete SAT solver
(respectively, G2-WSAT \cite{Li:05} and Satz-Rand \cite{Gomes:2000jar}) 
are combined to solve a sequence of random satisfiable
and unsatisfiable problems (benchmarks {\tt uf-*}, {\tt uu-*} from \cite{satlib}, $1899$ instances in total).
As the clauses-to-variable ratio is fixed in this benchmark, only the number of variables, ranging from $20$ to $250$,
was used as a problem feature $\x$.
Local search algorithms are more efficient on satisfiable instances,
but cannot prove unsatisfiability, so are doomed to run forever on unsatisfiable instances; while complete solvers
are guaranteed to terminate their execution on all
instances, as they can also prove unsatisfiability. 

\begin{figure*}[t]
\begin{minipage}{0.5\linewidth}
\includegraphics[width=\linewidth]{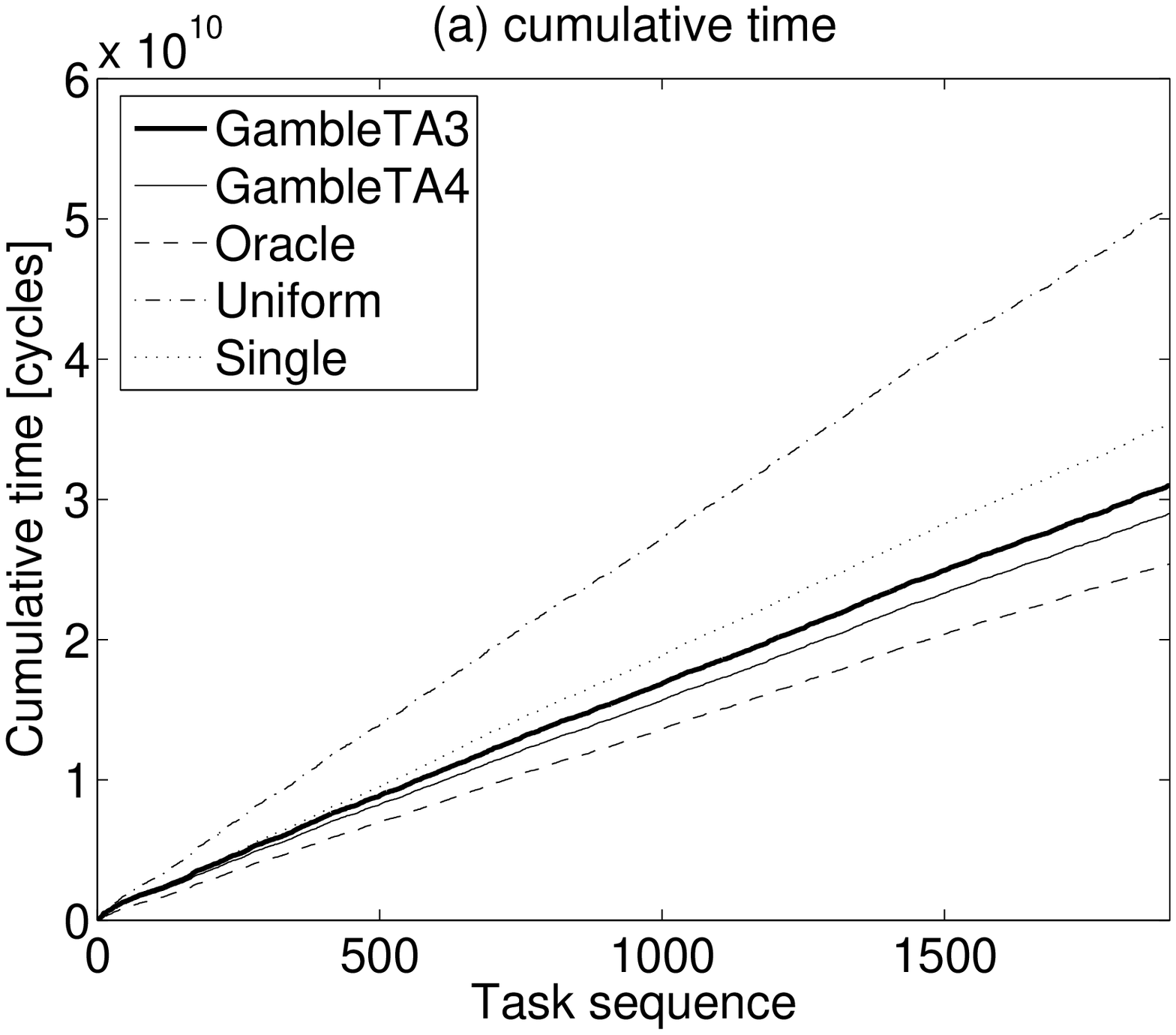}
\end{minipage}
\begin{minipage}{0.5\linewidth}
\includegraphics[width=\linewidth]{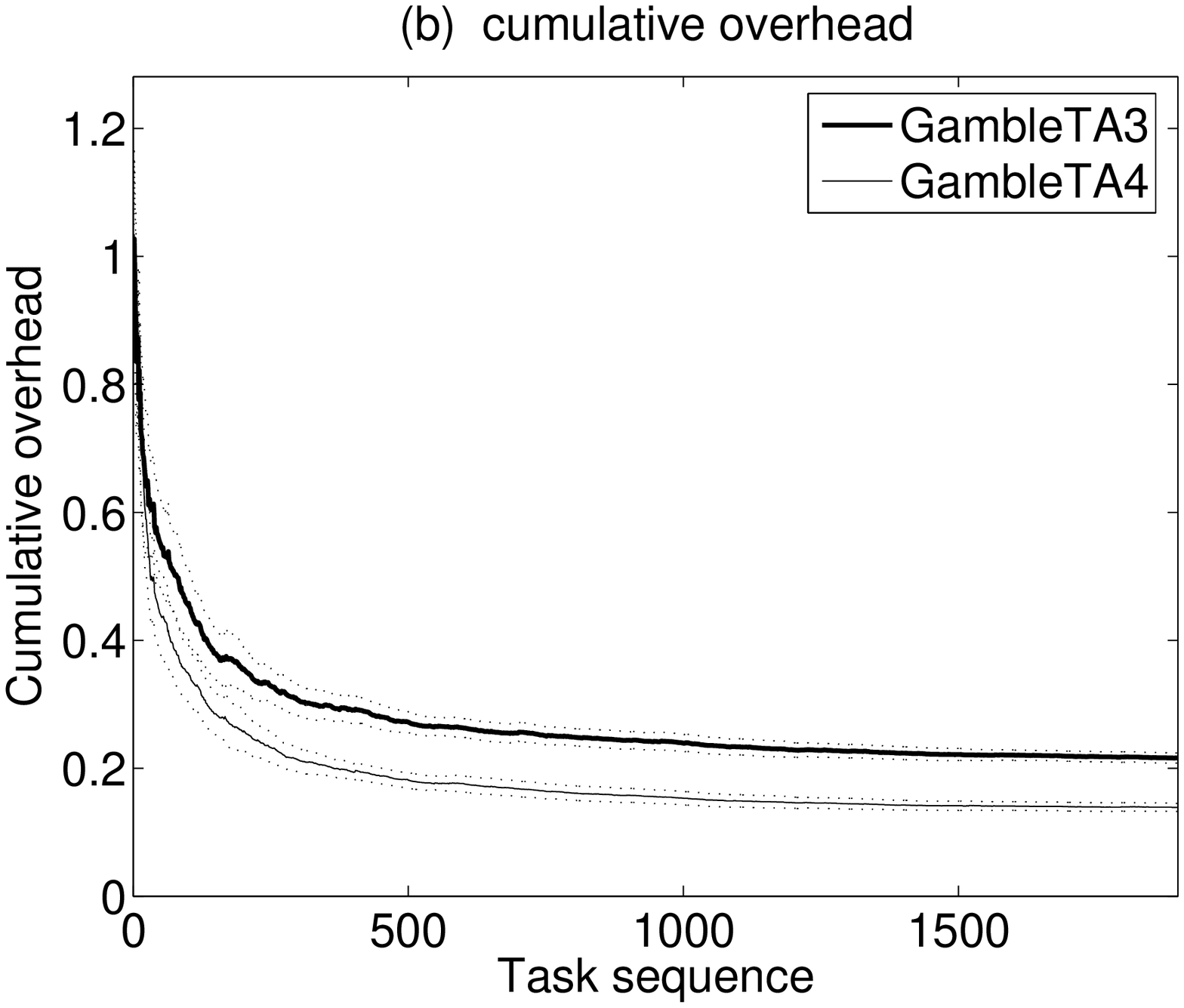}
\end{minipage}
\caption{\small (a): Cumulative time spent by \GambleTAThree\  and \GambleTAFour\ 
\cite{Gagliolo:06amai} on the SAT-UNSAT problem set ($10^{9}\approx 1$ min.). 
Upper $95\%$  confidence bounds on $\nruns$ runs, with random reordering of the problems.  \Oracle\ is the lower bound on 
performance. \Uniform\  is the ($0.5$,$0.5$) share. {\sc Satz-Rand} is the per-set best algorithm.
(b): The evolution of cumulative overhead, defined as $(\sum_j t_G(j)-\sum_j t_O(j))/\sum_j t_O(j)$,
 where $t_G$ is the performance of \GambleTA\  and $t_O$ is the performance of the oracle.
Dotted lines represent $95\%$ confidence bounds.
}
\label{gamblefig}
\end{figure*}

For the whole problem sequence, the overhead of 
\GambleTAThree\  (Algorithm \ref{gambleta}, using \ExpThreeLightA\  as the \BPS)
 over an ideal ``oracle'', which can predict
 and run only the fastest algorithm, is $22\%$.
 \GambleTAFour\ (from \cite{Gagliolo:06amai}, based on \ExpFour)
 seems to profit from the mixing of time allocation shares,
obtaining a  better $14\%$.
Satz-Rand alone can solve all the problems, but with an overhead of 
about $40\%$ w.r.t. the oracle, due to its poor performance
on satisfiable instances. Fig.~\ref{gamblefig} plots the evolution
of cumulative time, and cumulative overhead, along the problem sequence.

\section{Conclusions}
\label{conclusions}

We presented a bandit problem solver for loss games
with partial information and an unknown bound on losses.
The solver represents an ideal plug-in for our 
algorithm selection method \GambleTA, avoiding
 the need to set any additional parameter.
The choice of the algorithm set and time allocators
to use is still left to the user. Any existing selection
technique, including oblivious ones, can be included
in the set of $\nexps$ allocators, with an impact $O(\sqrt{\nexps})$
on the regret: the overall performance of \GambleTA\  will
converge to the one of the best time allocator.  
Preliminary experiments showed a degradation in performance
compared to the heuristic version presented in 
\cite{Gagliolo:06amai}, which requires to 
set in advance a maximum runtime, and cannot be provided
of a bound on regret.

According to \cite{cesa-bianchi08personal}, a bound for 
the original \ExpThreeLight\  can be proved for an adaptive 
$\eta_{\epochLbest}$ (\ref{eqetar}), in which the total number of trials 
$\ntrials$ is replaced by the current
trial $i$. This should allow for a potentially more efficient variation
 of \ExpThreeLightA,
in which \ExpThreeLight\  is not restarted at each 
epoch, and can retain the information
on past losses. 

One potential advantage of offline selection methods is that
the initial training phase can be easily parallelized, 
distributing the workload on a cluster of machines.
Ongoing research aims at extending \GambleTA\  to allocate
multiple CPUs in parallel, in order to obtain a fully
distributed algorithm selection framework \cite{Gagliolo:08dcai}.

{\bf Acknowledgments.}
We would like to thank Nicol\`{o} Cesa-Bianchi 
for contributing the proofs for \ExpThreeLight\  
and useful remarks on his work, 
and Faustino Gomez for his comments on a draft of 
this paper. This work was supported by the Hasler
foundation with grant n.~$2244$.

\bibliography{IDSIA-07-08}
\bibliographystyle{plain}

\section*{Appendix}
\renewcommand{\thesubsection}{A.\arabic{subsection}}

\subsection{Proof of Theorem \ref{thExp3Light}}

%\begin{proof}
The proof is trivially based on the regret for the original \ExpThreeLight,
with $\lmax=1$, which according to \cite[Theorem 5]{cesa-bianchi05improved} 
 (proof obtained from \cite{cesa-bianchi08personal}) can be evaluated
using the optimal values (\ref{eqetar}) for $\eta_{\epochLbest}$ :

\begin{eqnarray}
\label{eqExp3lightBound}
&&\Expval{\Lalg(\ntrials)}-\Loss^*(\ntrials) \leq \\
&&2\sqrt{2(\log \narms +\narms\log\ntrials)\narms(1+3\Loss^*(\ntrials))}\nonumber\\
&+& (2\narms+1)(1+\log_4(3\ntrials+1))\nonumber .
\end{eqnarray}

As we are playing the same game normalizing all losses with $\lmax$,
the following will hold for Alg. \ref{Exp3Light}:
\begin{eqnarray}
&&\frac{\Expval{\Lalg(\ntrials)}-\Loss^*(\ntrials)}{\lmax} \leq\\ 
&&2\sqrt{2(\log \narms +\narms\log\ntrials)\narms(1+3\Loss^*(\ntrials)/\lmax)}\\
&+&(2\narms+1)(1+\log_4(3\ntrials+1)) .
\end{eqnarray}

Multiplying both sides for $\lmax$ and rearranging produces (\ref{eqExp3lightBoundLmax}).\qed

\subsection{Proof of Theorem \ref{thExp3LightA}}

%\begin{proof}
This follows the proof technique employed in \cite[Theorem 4]{cesa-bianchi05improved}.
Be $i_{\epochlmax}$ the last trial of epoch $\epochlmax$, 
i.~e. the first trial at which a loss $\loss_{I(i)}(i)> 2^{\epochlmax}$ is observed.
Write cumulative losses during an epoch $\epochlmax$, \emph{excluding} the last trial $i_{\epochlmax}$, as
$\Loss^{(\epochlmax)}=\sum_{i=i_{\epochlmax-1}+1}^{i_{\epochlmax}-1} \loss(i)$,
and let  $\Loss^{*(\epochlmax)}=\min_j \sum_{i=i_{\epochlmax-1}+1}^{i_{\epochlmax}-1} \loss_j(i)$ indicate the optimal loss
for this subset of trials.
Be ${\lastepochlmax}=\epochlmax(\ntrials)$ the \emph{a priori} unknown epoch at the last trial.
In each epoch ${\epochlmax}$, the bound (\ref{eqExp3lightBoundLmax}) holds
with $\lmax_{\epochlmax}=2^{\epochlmax}$ for all trials except the last one $i_{\epochlmax}$,
so noting that $\log(\ntrials-i)\leq\log(\ntrials)$ we can write:

\begin{eqnarray}
\label{eqExp3lightAEpoch}
&&\Expval{\Lalg^{(\epochlmax)}}-\Loss^{*(\epochlmax)}\leq\\ 
%&&\sqrt{24\lmax_{\epochlmax}(\log \narms +\narms\log(\ntrials-i_{\epochlmax}+1))\narms\Loss^{*(\epochlmax)}}\\
%&+&2\lmax_{\epochlmax}\sqrt{2\lmax_{\epochlmax}(\log \narms +\narms\log(\ntrials-i_{\epochlmax}+1))\narms}\\
%&+&\lmax_{\epochlmax}(2\narms+1)(1+\log_4(3(\ntrials-i_{\epochlmax}+1)+1))\leq\\ 
&&2\sqrt{6\lmax_{\epochlmax}(\log \narms +\narms\log\ntrials)\narms\Loss^{*(\epochlmax)}}\nonumber\\
&+& \lmax_{\epochlmax}[2\sqrt{2\lmax_{\epochlmax}(\log \narms +\narms\log\ntrials)\narms}\nonumber\\
&+& (2\narms+1)(1+\log_4(3\ntrials+1))]\nonumber .
\end{eqnarray}
\noindent
 The loss for trial $i_{\epochlmax}$ can only be bound
by the next value of $\lmax_{\epochlmax}$, evaluated \emph{a posteriori}:
\begin{equation}
\label{eqExp3lightALast}
\Expval{\lalg(i_{\epochlmax})}-\loss^*(i_{\epochlmax})  
\leq \lmax_{\epochlmax+1} ,
\end{equation}
\noindent
where $\loss^{*}(i)=\min_j \loss_j(i)$ indicates the optimal loss at trial $i$.

Combining (\ref{eqExp3lightAEpoch},\ref{eqExp3lightALast}), 
%indicating the last 
%epoch as $\lastepochlmax$, 
and writing $i_{-1}=0$, $i_{\lastepochlmax}=\ntrials$,
 we obtain the regret for the whole game:\footnote{
Note that all cumulative losses are counted from trial $i_{\epochlmax-1}+1$ to
trial $i_{\epochlmax}-1$. If an epoch ends on its first trial, (\ref{eqExp3lightAEpoch})
is zero, and (\ref{eqExp3lightALast}) holds. Writing $i_{\lastepochlmax}=\ntrials$
implies the worst case hypothesis that the bound $\lmax_{\lastepochlmax}$ is exceeded on
the last trial. Epoch numbers $\epochlmax$ are increasing,
but not necessarily consecutive: in this case the terms related to the
missing epochs are $0$.
}
\begin{eqnarray}\nonumber
\label{eqExp3lightAMess}
&&\Expval{\Lalg(\ntrials)}-\sum_{\epochlmax=0}^{\lastepochlmax}\Loss^{*(\epochlmax)}
-\sum_{\epochlmax=0}^{\lastepochlmax} \loss^*(i_{\epochlmax})\\
&\leq&\sum_{\epochlmax=0}^{\lastepochlmax}\{2\sqrt{6\lmax_{\epochlmax}(\log \narms +\narms\log\ntrials)\narms\Loss^{*(\epochlmax)}}\nonumber\\
&+& \lmax_{\epochlmax}[2\sqrt{2\lmax_{\epochlmax}(\log \narms +\narms\log\ntrials)\narms}\nonumber\\
&+&(2\narms+1)(1+\log_4(3\ntrials+1))]\}\nonumber\\
&+&\sum_{\epochlmax=0}^{\lastepochlmax}\lmax_{\epochlmax+1}\nonumber .
\end{eqnarray}
%As the square root is a concave function, 
The first term on the right hand side 
of (\ref{eqExp3lightAMess}) can be bounded using Jensen's inequality 
\begin{equation} 
\sum_{\epochlmax=0}^{\lastepochlmax} \sqrt{a_{\epochlmax}}\leq \sqrt{(\lastepochlmax+1)\sum_{\epochlmax=0}^{\lastepochlmax} a_{\epochlmax}}, 
\end{equation}
\noindent
with 
\begin{eqnarray} 
a_{\epochlmax} &=& 24\lmax_{\epochlmax}(\log \narms +\narms\log\ntrials)\narms\Loss^{*(\epochlmax)}\\
&\leq& 24\lmax_{\lastepochlmax+1}(\log \narms +\narms\log\ntrials)\narms\Loss^{*(\epochlmax)}\nonumber.
\end{eqnarray}
\noindent
The other terms do not depend on the optimal losses $\Loss^{*(\epochlmax)}$, and can also
 be bounded noting that $\lmax_{\epochlmax}\leq\lmax_{\lastepochlmax+1}$.

%Note that the last term on the left hand side is $0$ if an epoch is started on the last trial $\ntrials$.

We now have to bound the number of epochs $\lastepochlmax$. This can 
be done %as in the proof of Theorem 4 from \cite{cesa-bianchi05improved}, 
noting that the maximum observed loss cannot be larger than the unknown, but finite, bound $\lmax$,
and that 
\begin{equation}
\label{eqEpochs}
{\lastepochlmax+1}=\ceil{\log_2 max_i \loss_{I(i)}(i)} \leq \ceil{\log_2\lmax} ,
\end{equation}
\noindent
which implies
\begin{equation}
\label{eqlastepochlmax}
\lmax_{\lastepochlmax+1}=2^{\lastepochlmax+1}\leq 2\lmax.
\end{equation}
\noindent
In this way we can bound the sum
\begin{equation}
\label{eqLmaxBound}
\sum_{\epochlmax=0}^{\lastepochlmax}\lmax_{\epochlmax+1}
\leq
\sum_{\epochlmax=0}^{\ceil{\log_2\lmax}} 2^{\epochlmax}
\leq
2^{1+\ceil{\log_2\lmax}}
\leq 4\lmax .
\end{equation}
\noindent
We conclude by noting that 
\begin{eqnarray}
\Loss^*(\ntrials)&=&min_j \Loss_j(\ntrials) \\ 
&\geq&\sum_{\epochlmax=0}^{\lastepochlmax}\Loss^{*(\epochlmax)} 
+\sum_{\epochlmax=0}^{\lastepochlmax} \loss^*(i_{\epochlmax})\geq
\sum_{\epochlmax=0}^{\lastepochlmax}\Loss^{*(\epochlmax)}\nonumber .
\end{eqnarray}
\noindent
Inequality (\ref{eqExp3lightAMess}) then becomes:
\begin{eqnarray}\nonumber
\label{eqExp3lightAMess2}
&&\Expval{\Lalg(\ntrials)}
-\Loss^{*}(\ntrials)\\
&\leq&2\sqrt{6(\lastepochlmax+1)\lmax_{\lastepochlmax+1}(\log \narms +\narms\log\ntrials)\narms\Loss^{*}(\ntrials)}\nonumber\\
&+&(\lastepochlmax+1)\lmax_{\lastepochlmax+1}[2\sqrt{2\lmax_{\lastepochlmax+1}(\log \narms +\narms\log\ntrials)\narms}\nonumber\\
&+&(2\narms+1)(1+\log_4(3\ntrials+1))]
+4\lmax\nonumber .
\end{eqnarray}
\noindent
Plugging in (\ref{eqEpochs}, \ref{eqlastepochlmax}) and rearranging we obtain (\ref{eqExp3lightABound}).\qed

%\end{proof}

\end{document}